%% file: main.tex
\documentclass[conference]{IEEEtran}
\IEEEoverridecommandlockouts
\usepackage{cite}
\usepackage{amsmath,amssymb,amsfonts}
\usepackage{graphicx}
\usepackage{textcomp}
\usepackage{xcolor}
\def\BibTeX{{\rm B\kern-.05em{\sc i\kern-.025em b}\kern-.08em
    T\kern-.1667em\lower.7ex\hbox{E}\kern-.125emX}}

\usepackage{graphicx} %
\usepackage{amsmath}
\usepackage{amssymb}

\usepackage{amsmath}
\usepackage{multirow}
\usepackage{tabularx}
\usepackage{pifont}
\usepackage{tikz}
\usepackage{xspace}
\usepackage{bm}

\usepackage{algorithm}
\usepackage[noend]{algpseudocode}

\usetikzlibrary{automata, positioning, shapes, arrows, shapes.geometric}

\input{commands}

\usepackage[capitalize]{cleveref}
\usepackage[shortcuts]{extdash}
\usepackage[all]{foreign}
\usepackage{glossaries}
\usepackage{arydshln}
\usepackage{fontawesome5}
\usepackage{xspace}

\usepackage{caption}
\usepackage{subcaption}

\usepackage{xcolor}
\definecolor{rorange}{HTML}{FFE6CC}
\definecolor{rblue}{HTML}{DAE8FC}
\definecolor{rgreen}{HTML}{D5E8D4}
\newcommand\coloredsquare[1]{\textcolor{#1}{\rule{1em}{1em}}}

\input{acronyms}

\usepackage{booktabs}
\usepackage[detect-all]{siunitx}

\begin{document}
\title{Assessing the Robustness of Intelligence-Driven Reinforcement Learning
\thanks{The work is partially supported by the European Office of Aerospace Research \& Development and the Air Force Office of Scientific Research under award number FA8655-22-1-7017 and by the US DEVCOM Army Research Laboratory (ARL) under Cooperative Agreement \#W911NF2220243. Any opinions, findings, and conclusions or recommendations expressed in this material are those of the author(s) and do not necessarily reflect the views of the United States government.}
}

\author{\IEEEauthorblockN{Lorenzo Nodari\IEEEauthorrefmark{1} and
Federico Cerutti\IEEEauthorrefmark{1}\IEEEauthorrefmark{2}}
\IEEEauthorblockA{
\IEEEauthorrefmark{1}University of Brescia, Italy}
\IEEEauthorblockA{\IEEEauthorrefmark{2}Cardiff University, UK}}

\maketitle

\begin{abstract}
    Robustness to noise is of utmost importance in reinforcement learning systems, particularly in military contexts where high stakes and uncertain environments prevail.
    Noise and uncertainty are inherent features of military operations, arising from factors such as incomplete information, adversarial actions, or unpredictable battlefield conditions. In RL, noise can critically impact decision-making, mission success, and the safety of personnel.
    Reward machines offer a powerful tool to express complex reward structures in RL tasks, enabling the design of tailored reinforcement signals that align with mission objectives. 
    This paper considers the problem of the robustness of intelligence-driven reinforcement learning based on reward machines.
    The preliminary results presented suggest the need for further research in evidential reasoning and learning to harden current state-of-the-art reinforcement learning approaches prior to being mission-critical-ready.
\end{abstract}

\begin{IEEEkeywords}
reinforcement learning, intelligence gathering, artificial intelligence
\end{IEEEkeywords}

\section{Introduction}
\input{sections/introduction}

\section{Background}
\input{sections/background}

\section{Methodology}
\input{sections/methodology}

\input{sections/results}

\section{Conclusions}
\input{sections/conclusions}

\bibliographystyle{IEEEtran}

\bibliography{biblio,added_biblio}

\end{document}

%% file: commands.tex
\newcommand{\tuple}[1]{\ensuremath{\left\langle #1 \right\rangle}}

\newcommand{\expect}{\ensuremath{\mathbb{E}}}

\newcommand{\pomdp}{\mathcal{P_O}} %
\newcommand{\events}{\mathcal{P}} %
\newcommand{\lf}{L} %

\newcommand{\rmname}{M}
\newcommand{\rmstates}{\mathcal{U}}
\newcommand{\propositions}{\mathcal{P}}
\newcommand{\rmdeltau}{\delta_u}
\newcommand{\rmdeltar}{\delta_r}
\newcommand{\rmstate}{u}
\newcommand{\rminitstate}{\rmstate_0}
\newcommand{\rmstatefinal}{\rmstate_A}

\newcommand{\mdpdisc}{\gamma}

\newcommand{\proplabel}{\mathcal{L}}

\newcommand{\gotclubsuit}{\textcolor{red}{\clubsuit}}
\newcommand{\gotspadesuit}{\textcolor{red}{\spadesuit}}
\newcommand{\gotzolenge}{\textcolor{red}{\blacklozenge}}

\newcommand{\qfunc}{q}

\algblockdefx{Static}{EndStatic}{\textbf{static variables:}}{}

%% file: acronyms.tex
\newacronym{mdp}{MDP}{Markov Decision Process}
\newacronym{pomdp}{POMDP}{Partially Observable Markov Decision Process}
\newacronym{uav}{UAV}{unmanned aerial vehicles}
\newacronym{poi}{PoI}{Person of Interest}
\newacronym{humint}{HUMINT}{Human Intelligence}

%% file: sections/introduction.tex
\label{sec:introduction}

Suppose a \gls{poi} is allegedly hiding in a compound whose blueprint has been leaked by an informant as part of an \gls{humint} gathering process. A new intelligence gathering mission planning starts with the goal of confirming beyond any doubt the presence of the \gls{poi}.
\gls{poi}'s acolytes can enter the compound from an opening and announce them in an interphone. Based on their identity and the history of interactions, they then go into one of the two rooms next to the one with the interphone, where the \gls{poi} reaches them from behind an armoured door.
Through the informant, enough samples of an acolyte's voice have been recorded, enough for building a text-to-speech synthesiser. It is, however, impossible to predict which room the acolyte should use to meet with the \gls{poi}.

The compound's location deep into enemy territory makes it impossible to deploy human forces, and the thickness of the walls impacts the remote control of micro \glspl{uav}. 
An option is to use a fully autonomous micro \gls{uav} \--- equipped with a loudspeaker and a camera with face recognition software \--- which can be trained on a replica of the three rooms based on the leaked blueprint.

\newcommand{\rzero}{\faPhone\xspace}
\newcommand{\rone}{\faChevronCircleRight\xspace}
\newcommand{\rtwo}{\faChevronCircleLeft\xspace}
\newcommand{\reward}{\faDollarSign\xspace}
\newcommand{\failreward}{\faFrown[regular]\xspace}
\newcommand{\zeroreward}{}
\newcommand{\anything}{\faAsterisk\xspace}
\newcommand{\somebody}{\faPortrait\xspace}
\newcommand{\found}{\faWifi\xspace}
\newcommand{\talked}{\faCommentDots\xspace}

\begin{figure}[t]
    \centering

\resizebox{0.9\columnwidth}{!}{
\begin{tikzpicture}[node distance=3cm,on grid,
  every initial by arrow/.style={->, >=stealth}, initial text={}]
  \node[state,initial] (u_0) at (0,0) {$u_0$};
  \node[state]         (u_1) at (0,2) {$u_1$};
  \node[state]         (u_2) at (3.5,2) {$u_2$};
  \node[state]         (u_3) at (-3.5,2) {$u_3$};
  \node[state]         (u_4) at (0, 4.5) {$u_4$};

  \path[->] (u_0) edge [->, >=stealth,loop below] node [right]{$\tuple{\text{\anything},\text{\zeroreward}}$} ();
  \path[->] (u_1) edge [->, >=stealth,loop above] node {$\tuple{\text{\anything},\text{\zeroreward}}$} ();
  \path[->] (u_2) edge [->, >=stealth,loop below] node {$\tuple{\text{\anything},\text{\zeroreward}}$} ();
  \path[->] (u_3) edge [->, >=stealth,loop below] node {$\tuple{\text{\anything},\text{\zeroreward}}$} ();
  
  \path[->, >=stealth] (u_0) edge node [right] {$\tuple{\text{\rzero\talked},\text{\zeroreward}}$} (u_1);
  \path[->, >=stealth] (u_1) edge node [above] {$\tuple{\left[\text{\rone\somebody} \mid \text{\rtwo}\right],\text{\zeroreward}}$}(u_2);
  \path[->, >=stealth] (u_1) edge node [above] {$\tuple{\left[\text{\rtwo\somebody} \mid \text{\rone}\right],\text{\zeroreward}}$} (u_3);
  \path[->, >=stealth] (u_2) edge[bend right] node [right] {$\tuple{\text{\found},\text{\reward}}$} (u_4);
  \path[->, >=stealth] (u_3) edge[bend left] node [left] {$\tuple{\text{\found},\text{\reward}}$} (u_4);

\end{tikzpicture}}

    \caption{High-level description of the autonomous micro \gls{uav} behaviour based on the gathered \gls{humint} represented as a finite-state automaton. Each state-transitions is labelled with a pair, where the first element is the set of high-level observations in the environment, and the second is the reward the agent receives in performing the transition: when empty, the reward is 0. \anything is a shortcut for \emph{any other input}.}
    \label{fig:humintrm}
\end{figure}
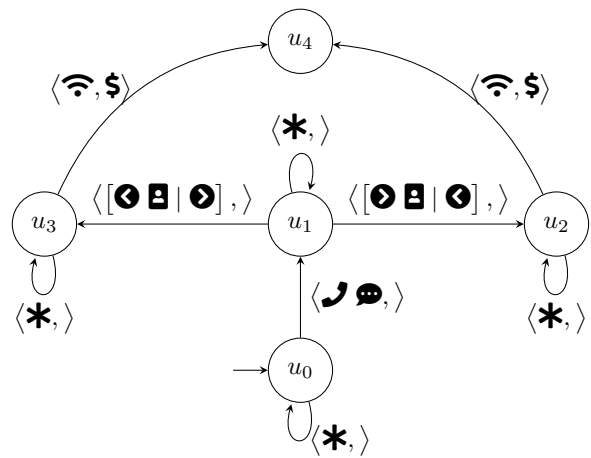

Based on the gathered \gls{humint}, the high-level behaviour of the autonomous micro \gls{uav} is simple enough that can be captured by the finite-state automaton depicted in \cref{fig:humintrm}. Waking up in the state $u_0$, the autonomous agent will identify the entrance of the room with the interphone \rzero and impersonate the acolyte's voice \talked. This triggers the transition into the state $u_1$, in which the autonomous agent will have to explore the two adjacent rooms, one on the right \rone and one on the left \rtwo of the one with the interphone. Once the camera's software notices a human figure, it runs the face recognition software to identify the \gls{poi} \somebody. At this point, the agent leaves the compound to transmit a confirmation signal back to base \found, when it finally receives a reward \reward \--- all other transactions have null reward \--- and reaches the final state $u_4$.

While our motivating scenario aims at being a simple toy example, real-world settings may be characterized by a number of additional difficulties, with a highly complex compound structure and the need for the agent's ability to avoid detection being just two notable examples. Therefore, in general, writing a program which implements the proper high-level behaviour to carry out the mission is non-trivial: the micro \gls{uav} can learn it from interactions with the environment, using reinforcement learning \cite{SuttonB18}. To make use of the intelligence gathered and encoded in \cref{fig:humintrm}, a recent proposal for reinforcement learning is used, namely Reward Machines \cite{rm_2022}, which is summarised in \cref{sec:back_rl}. A reward machine can use the description provided in \cref{fig:humintrm} and use it to train specific behaviours (policies) for each of the states in the automaton.

The research community still did not provide a convincing analysis of the robustness of reward machines, which is paramount when the stakes are high; \eg the autonomous micro \gls{uav} is a costly research prototype, and the intelligence gathering mission is necessary for identifying suitable courses of action. In light of this, \Cref{sec:metodology} details a methodology for assessing the robustness of reward machines, looking at the specifics of our approach. \Cref{sec:results} illustrates the preliminary results gathered so far: they support the intuition that more robust and uncertainty-aware \cite{cerutti2022handling,cerutti2022evidential} approaches are needed for ensuring robust exploitation of learned policies.

%% file: sections/background.tex
\label{sec:back_rl}
In reinforcement learning, an agent interacts with an unknown environment, usually modelled as a \gls{pomdp}, $\pomdp = \langle S, O, A, r, p, \omega, \gamma \rangle$, where $S$ is the set of \emph{states} of the environment, $O$ is the set of \emph{observations}, $A$ is the set of agent \emph{actions}, $r: S \times A \times S \rightarrow \mathbb{R}$ is the \emph{reward function}, $p(s' | s, a)$ is the \emph{environment transition model}, $\omega(o | s)$ is the \emph{observation probability distribution} and $\gamma$ is the \emph{discount factor}.

At a given time step $t$, the agent \--- being in the state $s_t \in S$ \--- selects an action $a_t$ based upon a probability distribution or \emph{policy} $\pi(\cdot \mid s_t)$. Further to executing $a_t$, the agent enters a new state $s_{t+1} \sim p(\cdot|s_t,a_t)$ and receives an immediate reward $r(s_t,a_t,s_{t+1})$ from the environment. 
An \emph{optimal policy} $\pi^*$ maximises the expected discounted return 
\begin{equation}
    G_t = \expect_{\pi}{\left[\sum_{k=0}^{\infty}\gamma^kr_{t+k} \mid s_t=s\right]}
\end{equation}
for any state $s \in S$ and time step $t$.

A \textbf{reward machine} \cite{rm_2022} is a finite-state machine that receives abstracted descriptions of the environment as inputs, and it outputs reward functions. It is defined over a set of propositional symbols $\mathcal{P}$. Intuitively, $\mathcal{P}$ is a set of relevant high-level events from the environment that the agent can detect. For instance, concerning the example introduced in \cref{sec:introduction}, each of the symbols triggering transitions \--- \eg \rzero, \found, \ldots \--- are elements of such set $\mathcal{P}$.

Formally, a reward machine is a tuple $\rmname =\langle\rmstates,\propositions, \rminitstate, \rmstatefinal, \rmdeltau, \rmdeltar\rangle$ where $\rmstates$ is a set of states; $\propositions$ is a set of propositions; $\rminitstate\in\rmstates$ is the initial state; $\rmstatefinal\in\rmstates$ is the final state; $\rmdeltau:\rmstates \times 2^\propositions \to \rmstates$ is a state-transition function such that $\rmdeltau(u,\proplabel)$ is the state that results from observing label $\proplabel\in 2^\propositions$ in state $u\in\rmstates$; and $\rmdeltar:\rmstates \times \rmstates \to \mathbb{R}$ is a reward-transition function such that $\rmdeltar(u,u')$ is the reward obtained for transitioning from state $u\in \rmstates$ to $u'\in\rmstates$. We assume that (i)~there are no outgoing transitions from $\rmstatefinal$, and (ii)~$\rmdeltar(\rmstate,\rmstate')=1$ if $\rmstate'=\rmstatefinal$ and 0 otherwise.

A reward machine can be used by a reinforcement learning agent as a high-level, structured representation of the current state of the environment: each state $u_i \in \rmstates$ can be thought of as a synthesis of various properties of the world that the agent can use to guide its actions. Thus, under this interpretation, each propositional symbol $p \in \events$ represents an atomic high-level event that, when verified, can trigger a state change in the reward machine.

To use a reward machine, an agent needs access to a \textbf{labelling function} that allows it to detect the high-level events that trigger the transitions between different reward machine states, thus allowing it to properly keep track of the relevant high-level state of the environment. More specifically, a labelling function can be defined as a function $\lf : O \times A \times O \rightarrow 2^\events$. Given the current observation $o \in O$, the action taken by the agent $a \in A$ and the resulting observation $o' \in O$, $\lf(o, a, o')$ is the set of high-level events that currently hold in the environment.

The Q-learning for reward machines (QRM) algorithm~\cite{rm_2022} exploits the task structure modelled through reward machines. QRM learns a Q-function $\qfunc_\rmstate$ for each state $\rmstate\in\rmstates$ in the reward machine. Given an experience tuple $\langle s,a,s'\rangle$, a Q-function $\qfunc_\rmstate$ is updated as follows:
\begin{equation}
\begin{split}
    \qfunc_\rmstate(s,a) = & \qfunc_\rmstate(s,a)+\alpha\left(\rmdeltar(\rmstate,\rmstate') \right.+ \\
                            & \left. \mdpdisc\max_{a'}\qfunc_{\rmstate'}(s',a') - \qfunc_\rmstate(s,a) \right),
\end{split}
\end{equation}
where $\rmstate'=\rmdeltau(\rmstate,\lf(s, a, s'))$. All Q-functions (or a subset of them) are updated at each step using the same experience tuple in a counterfactual manner.
In the tabular case, QRM is guaranteed to converge to an optimal policy in the limit.

%% file: sections/methodology.tex
\label{sec:metodology}

Since the labelling function is the only way an agent has to access the high-level state of the environment, its proper functioning is of paramount importance in assuring the correctness of its decision-making process. For this reason, our work aims to analyse the potential impact, from the agent's performance point of view, arising from an agent's reliance on incorrect labelling function outputs. Specifically, we consider the effects arising when trained agents are exposed to varying amounts of unforeseen noise. In other words, we first train our agents in noiseless environments until they reach optimal levels of performance, and then we quantify the decrease in their proficiency associated with different amounts of noise in the labelling function outputs. 

In this preliminary analysis, we limit ourselves to studying the effect of \emph{random} noise from the environment. In particular, we consider the case where each labelling function output is altered, with a given probability --- the \textbf{noise level} --- by tampering with a single, randomly chosen observation included in the original output. Each tampering can consist of either the removal or substitution of the original observation with a randomly chosen one. Despite this being one of the most basic cases to consider, it can effectively serve as a proxy for a large variety of practical situations, from sensor failure to adversarial attacks in a contested environment. Algorithm \ref{alg:noise_injection} provides a pseudo-code implementation of the noise-injection procedure followed during our experiments.

\begin{algorithm}
\caption{Pseudocode for the noise injection procedure followed during our experiments for tampering with a labelling function output $\proplabel \in 2^\propositions$, under a given noise level $k \in [0,1]$}
\label{alg:noise_injection}

    \begin{algorithmic}
        \Function{Labelling-Noise}{$\proplabel, k$}
        \Static
            \State $\propositions$: the set of all possible high-level events
        \EndStatic

        \State $tamper? \gets \Call{Random}{0,1} < k$
        \If{not $tamper?$}
            \State \textbf{return} $\proplabel$
        \Else
        
            \State $e \gets \Call{Random-Choice}{\proplabel}$ \Comment Target event
            \State $\Tilde{e} \gets \Call{Random-Choice}{\propositions}$ \Comment Substitute event

            \If{$e = \Tilde{e}$}
                \State $\Tilde{\proplabel} \gets \proplabel - \{e\}$ \Comment Remove target
            \Else
                \State $\Tilde{\proplabel} \gets \proplabel - \{e\} \cup \{\Tilde{e}\}$ \Comment Perform substitution
            \EndIf
            \State \textbf{return} $\Tilde{\proplabel}$
        \EndIf

        \EndFunction
    \end{algorithmic}
    
\end{algorithm}

\begin{figure*}
     \centering
     \begin{subfigure}[b]{0.35\textwidth}
         \centering
         \includegraphics[width=\textwidth]{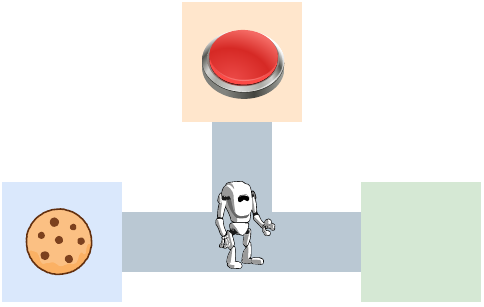}
         \caption{CookieWorld}
         \label{fig:cookieworld}
     \end{subfigure}
     \qquad
     \begin{subfigure}[b]{0.35\textwidth}
         \centering
         \includegraphics[width=\textwidth]{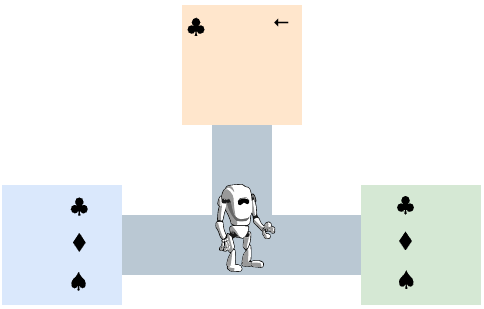}
         \caption{SymbolWorld}
         \label{fig:symbolworld}
     \end{subfigure}
     \caption{Partially observable environments: the agent can observe only the content of a single room.}
      \label{fig:domains}
\end{figure*}

We performed an experimental analysis \--- \cf \cref{sec:results} \--- based on two different grid-world environments, CookieWorld and SymbolWorld, both of them introduced in \cite{lrm_2019} and depicted in \cref{fig:domains}.

The \textbf{CookieWorld} domain \--- \cf \cref{fig:cookieworld} \--- is analogous to our motivating example discussed in \cref{sec:introduction}. An agent can access three rooms \--- orange, green, and blue \--- connected by a hallway. The agent can move in the four cardinal directions. There is a button in the orange room that, when pressed, causes a cookie to randomly appear in the green or blue room. The agent receives a reward of $\num{+1}$ for reaching and eating the cookie. Pressing the button before reaching a cookie will remove the existing cookie and cause a new cookie to appear randomly. There is no cookie at the beginning of the episode. This domain is partially observable since the agent can only see what it is in the room that it currently occupies.

As per the previous domain, in the \textbf{SymbolWorld} domain \--- \cf \cref{fig:symbolworld} \---  an agent can access three rooms \--- orange, green, and blue \--- connected by a hallway.
Differently than before, this domain has three symbols $\clubsuit$, $\spadesuit$, and $\blacklozenge$ in the blue and green rooms. At the beginning of an episode, one symbol from $\{\clubsuit, \spadesuit, \blacklozenge\}$ and possibly a right or left arrow are randomly placed at the orange room. Intuitively, that symbol and arrow will tell the agent where to go. For example, $\clubsuit$ and $\rightarrow$ tell the agent to go to $\clubsuit$ in the east room. If there is no arrow, the agent can go to the target symbol in either room. An episode ends when the agent reaches any symbol in the blue or green room, at which point it receives a reward of $\num{+1}$ if it reaches the correct symbol and $\num{-1}$ otherwise. All other steps in the environment provide no reward.

\newcommand{\rbutton}{\coloredsquare{rorange}}
\newcommand{\rblue}{\coloredsquare{rblue}}
\newcommand{\rgreen}{\coloredsquare{rgreen}}
\newcommand{\pressed}{\faHandPointDown}
\newcommand{\eatcookie}{\faCookieBite}
\newcommand{\seecookie}{\faCookie}

\begin{figure}[t]
    \centering

\resizebox{0.9\columnwidth}{!}{
\begin{tikzpicture}[node distance=3cm,on grid,
  every initial by arrow/.style={->, >=stealth}, initial text={}]
  \node[state,initial] (u_0) at (0,0) {$u_0$};
  \node[state]         (u_1) at (0,2) {$u_1$};
  \node[state]         (u_2) at (3.5,2) {$u_2$};
  \node[state]         (u_3) at (-3.5,2) {$u_3$};
  \node[state]         (u_4) at (0, 5) {$u_4$};

  \path[->] (u_0) edge [->, >=stealth,loop below] node [right]{$\tuple{\text{\anything},\text{\zeroreward}}$} ();
  \path[->] (u_1) edge [->, >=stealth,loop above] node {$\tuple{\text{\anything},\text{\zeroreward}}$} ();
  \path[->] (u_2) edge [->, >=stealth,loop below] node {$\tuple{\text{\anything},\text{\zeroreward}}$} ();
  \path[->] (u_3) edge [->, >=stealth,loop below] node {$\tuple{\text{\anything},\text{\zeroreward}}$} ();
  
  \path[->, >=stealth] (u_0) edge node [right] {$\tuple{\text{\rbutton\pressed},\text{\zeroreward}}$} (u_1);
  \path[->, >=stealth] (u_1) edge[bend left] node [above] {$\tuple{\left[\text{\rblue\seecookie} \mid \text{\rgreen}\right],\text{\zeroreward}}$}(u_2);
  \path[->, >=stealth] (u_1) edge[bend right] node [above] {$\tuple{\left[\text{\rgreen\seecookie} \mid \text{\rblue}\right],\text{\zeroreward}}$} (u_3);
  \path[->, >=stealth] (u_2) edge[bend right] node [right] {$\tuple{\text{\eatcookie},\text{\reward}}$} (u_4);
  \path[->, >=stealth] (u_3) edge[bend left] node [left] {$\tuple{\text{\eatcookie},\text{\reward}}$} (u_4);
  \path[->, >=stealth] (u_2) edge[bend left] node [above] {$\tuple{\text{\rbutton\pressed},\text{\zeroreward}}$} (u_1);
  \path[->, >=stealth] (u_3) edge[bend right] node [above] {$\tuple{\text{\rbutton\pressed},\text{\zeroreward}}$} (u_1);

\end{tikzpicture}}

    \caption{High-level description of the reward machine considered for the CookieWorld domain, \cf \cref{fig:cookieworld}. Each state-transitions is labelled with a pair, where the first element is the set of high-level observations in the environment, and the second is the reward the agent receives in performing the transition: when empty, the reward is 0. \anything is a shortcut for \emph{any other input}. This figure is based upon \cite[Fig. 2c]{lrm_2019}.}
    \label{fig:cookieworldrm}
\end{figure}

\begin{figure}[t]
    \centering

\resizebox{0.9\columnwidth}{!}{
\begin{tikzpicture}[node distance=3cm,on grid,
  every initial by arrow/.style={->, >=stealth}, initial text={}]
  \node[state,initial] (u_0) at (-3.5,0) {$u_0$};
  \node[state]         (u_1) at (0,4) {$u_1$};
  \node[state]         (u_4) at (0,0) {$u_4$};
  \node[state]         (u_9) at (0,-4) {$u_9$}; 
  \node[state]         (u_10) at (3.5,0) {$u_{10}$};
  \node[rotate=90] (dots1) at (0,2) {$\cdots\cdots$};
  \node[rotate=90] (dots2) at (0,-2.5) {$\cdots\cdots$};

  \path[->] (u_0) edge [->, >=stealth,loop below] node [below]{$\tuple{\text{\anything},\text{\zeroreward}}$} ();
  \path[->] (u_1) edge [->, >=stealth,loop above] node [above]{$\tuple{\text{\anything},\text{\zeroreward}}$} ();
  \path[->] (u_9) edge [->, >=stealth,loop below] node [below]{$\tuple{\text{\anything},\text{\zeroreward}}$} ();
  \path[->] (u_4) edge [->, >=stealth,loop below] node [below]{$\tuple{\text{\anything},\text{\zeroreward}}$} ();
  
  \path[->, >=stealth] (u_0) edge node [above left] {$\tuple{\text{\rbutton}\clubsuit\rightarrow,\text{\zeroreward}}$} (u_1);
  \path[->, >=stealth] (u_0) edge node [above] {$\tuple{\text{\rbutton}\clubsuit,\text{\zeroreward}}$} (u_4);
  \path[->, >=stealth] (u_0) edge node [below left] {$\tuple{\text{\rbutton}\clubsuit\leftarrow,\text{\zeroreward}}$} (u_9);
  \path[->, >=stealth] (u_1) edge node [above right] {$\tuple{\text{\rgreen}\gotclubsuit,\text{\reward}}$} (u_10);
  \path[->, >=stealth] (u_1) edge [bend left=90] node [above right, near start] {$\tuple{\text{\rblue}\gotclubsuit,\text{\failreward}}$} node [above right] {$\tuple{\gotspadesuit,\text{\failreward}}$} node [above right, near end] {$\tuple{\gotzolenge,\text{\failreward}}$} (u_10);
  \path[->, >=stealth] (u_4) edge node [above] {$\tuple{\text{\rgreen}\gotclubsuit,\text{\reward}}$} (u_10);
  \path[->, >=stealth] (u_4) edge node [below] {$\tuple{\text{\rblue}\gotclubsuit,\text{\reward}}$} (u_10);
  \path[->, >=stealth] (u_9) edge node [below right] {$\tuple{\text{\rblue}\gotclubsuit,\text{\reward}}$} (u_10);

\end{tikzpicture}}

    \caption{Partial high-level description of the reward machine considered for the SymbolWorld domain --- \cf \cref{fig:symbolworld} --- depicting only the states and transitions relating to the task of reaching a $\clubsuit$ symbol. Each state-transitions is labelled with a pair, where the first element is the set of high-level observations in the environment, and the second is the reward the agent receives in performing the transition: when empty, the reward is 0. \anything is a shortcut for \emph{any other input}. This figure is based upon \cite[Fig. 2c]{lrm_2019}. Note that, for ease of depiction, the transitions associated with agent failure are shown only for the $u_1, u_{10}$ pair of states, but can be easily deduced for any other pair of states.}
    \label{fig:symbolworldrm}
\end{figure}

To test our approach, we trained 10 agents for each environment using the QRM algorithm. All the agents were supplied with hand-crafted perfect reward machines for their corresponding environment. For instance, \cref{fig:cookieworldrm} depicts the reward machine considered for the CookieWorld domain. In there, any transition is labelled with the high-level observations coming from the environment and the associated reward $\rmdeltar(u,u')$, \cf \cref{sec:back_rl}. 
In particular, the high-level observations are:
\begin{itemize}
    \item \rbutton, being in the orange room;
    \item \rblue, being in the blue room;
    \item \rgreen, being in the green room;
    \item \pressed, having pressed the button;
    \item \seecookie, being in the same room with the cookie;
    \item \eatcookie, eating the cookie.
\end{itemize}
It is clearly very similar to the reward machine considered in our motivating example \--- \cf \cref{fig:humintrm} \--- with the notable difference that in the CookieWorld the agent can press the button before reaching a cookie: this removes the existing cookie and causes a new cookie to appear randomly.
Similarly, \cref{fig:symbolworldrm} presents a partial view of the reward machine considered for the SymbolWorld domain, limited, for ease of depiction, to some of the states and transitions relating to the task of reaching a $\clubsuit$ symbol. In addition to the events indicating the current room, in this case, the high-level observations are:
\begin{itemize}
    \item $\clubsuit, \spadesuit, \blacklozenge$, indicating that the agent is seeing the corresponding symbol;
    \item $\gotclubsuit, \gotspadesuit, \gotzolenge$, indicating that the agent has collected the corresponding symbol;
    \item $\rightarrow$, indicating that the agent should collect the target symbol in the green room;
    \item $\leftarrow$, indicating that the agent should collect the target symbol in the blue room.
\end{itemize}
Moreover, as the agent can now \emph{fail} its task by either collecting the wrong symbol or collecting the correct one in the wrong room, \failreward is used to indicate the reward associated with agent failure.

As per performance metrics, in this preliminary work, we limit ourselves to two:
\begin{enumerate}
    \item average success rate, a statistic showing the expected probability of success for any episode;
    \item average steps to success, a statistics showing the expected number of steps an agent will have to perform to achieve the reward.
\end{enumerate}

Our \textbf{main experimental hypothesis} is the following: an increase in the noise in the observations of the world should lead to a decrease in the average success rate and an increase in the average steps to success.

%% file: sections/results.tex
\begin{table*}[t]
\centering
\caption{Robustness against random noise for the CookieWorld domain (\subref{tab:rescookie}) and the SymbolWorld domain (\subref{tab:ressymbol}).}
\label{tab:resnoisefull}
\renewcommand{\arraystretch}{1.5}
    \begin{subtable}[h]{\textwidth}
    \centering
    \caption{}
    \label{tab:rescookie}
    \begin{tabular}{@{} S[table-format=3.2,table-number-alignment=right] S[table-format=3.2,table-number-alignment=right] S[table-format=3.2,table-number-alignment=right] S[table-format=3.2,table-number-alignment=right] S[table-format=3.2,table-number-alignment=right]  @{}}
    \toprule
    \textbf{Noise level (\%)} & \textbf{Avg. Success Rate (\%)} & \textbf{Avg. Steps to Success} & \textbf{Avg. Steps to Failure} & \textbf{Avg. Failure Reward} \\
    \midrule
    1.00                             & 98.28           & 35.96     & 500 & 0                     \\
    \hdashline
    5.00                             & 92.02           & 36.54     & 500 & 0                 \\
    \hdashline
    10.00                           & 84.24           & 37.23      & 500 & 0                 \\
    \hdashline
    20.00                            & 71.86           & 39.34    & 500 & 0                  \\
    \hdashline
    30.00                            & 61.35           & 41.35    & 500 & 0                   \\
    \hdashline
    40.00                            & 53.13           & 42.94    & 500 & 0                   \\
    \hdashline
    50.00                            & 45.18           & 45.02    & 500 & 0        \\
    \bottomrule
    \end{tabular}
    \end{subtable}

    \vspace{1em}

    \begin{subtable}[h]{\textwidth}
    \centering
    \caption{}
    \label{tab:ressymbol}
    \begin{tabular}{@{} S[table-format=3.2,table-number-alignment=right] S[table-format=3.2,table-number-alignment=right] S[table-format=3.2,table-number-alignment=right] S[table-format=3.2,table-number-alignment=right] S[table-format=3.2,table-number-alignment=right] @{}}
    \toprule
    \textbf{Noise level (\%)} & \textbf{Avg. Success Rate (\%)} & \textbf{Avg. Steps to Success} & \textbf{Avg. Steps to Failure} & \textbf{Avg. Failure Reward} \\
    \midrule
    1.00                          & 99.91           & 18.64      &18.83 &-1           \\
    \hdashline
    5.00                           & 99.44           & 18.87     &18.85 &-1            \\
    \hdashline
    10.00                         & 99.30           & 19.13      &19.14 &-1           \\
    \hdashline
    20.00                         & 97.97           & 19.75      &19.58 &-1           \\
    \hdashline
    30.00                         & 96.58           & 20.49      &20.62 &-1           \\
    \hdashline
    40.00                         & 95.35           & 21.37      &21.48 &-1            \\
    \hdashline
    50.00                         & 92.87           & 22.35     &22.65 &-1           \\
    \bottomrule
    \end{tabular}
    \end{subtable}
\end{table*}

\begin{table}[t]
    \centering
    \renewcommand{\arraystretch}{1.5}
    \caption{Baseline metrics for agents trained without noise.}
    \label{tab:baseline}
    \begin{tabular}{@{} l S[table-format=3.2,table-number-alignment=right] S[table-format=3.2,table-number-alignment=right]}
         \toprule
         \textbf{Domain} & \textbf{Avg. Success Rate (\%)} & \textbf{Avg. Steps to Success}\\
         \midrule
         CookieWorld & 100.00 & 35.77\\
         \hdashline
         SymbolWorld & 100.00 & 18.59\\
         \bottomrule
    \end{tabular}
\end{table}

\section{Results}

\label{sec:results}

After being trained, the performance of each agent was first assessed in a baseline session, where the agent was free to act in its environment without any external intervention, \emph{i.e.}: a noise level of 0\%. Each agent was tested on 1000 different episodes, with a time limit of 500 steps: if, after this amount of time the agent still hadn't achieved its task or, in the case of the SymbolWorld domain, failed it, the episode was terminated and the agent was given a null reward. \Cref{tab:baseline} summarises the performance of these agents: since the reward machine is perfectly designed to match the problem, it is not a surprise that the average success rate is 100.00\% for both environments.

We then injected random noise, as discussed in \cref{sec:metodology}. \cref{tab:resnoisefull} summarises the experimental results gathered, both for the CookieWorld \--- \cref{tab:rescookie} \--- and for the SymbolWorld \--- \cref{tab:ressymbol}. By visual inspection, we can confirm that there is evidence in support of our experimental hypotheses.

We can, however, notice substantial differences between the two domains. CookieWorld seems to be quite sensitive to the noise: \eg considering a 30\% observation noise leads to a decrease of the average success rate of nearly 40\%. SymbolWorld, instead, seems to be more resilient: \eg a 30\% noise leads to a decrease of the average success rate of less than 4\%.
This is in part explained by the presence of many more possible high-level observations in the SymbolWorld domain, as well as its more deterministic structure. However, when observing the failure statistics we notice that SymbolWorld failures were always caused by the agents reaching the wrong symbol, thus highlighting how carefully timed noise can lead to a complete misunderstanding of the task, a critical issue in practical deployments. To clarify this intuition, consider the task of reaching the $\clubsuit$ in the left room, indicated by the high-level observation $\proplabel= \tuple{\text{\rbutton},\clubsuit,\leftarrow}$, as in \cref{fig:symbolworldrm}. If the agent receives this observation while in state $u_0$, it successfully transitions to state $u_1$, thus guaranteeing the correct understanding of its task. However, if the random noise manages to alter exactly the first occurrence of such observation, substituting it, for instance, with the tampered $\Tilde{\proplabel} = \tuple{\text{\rbutton}, \spadesuit, \leftarrow}$, the agent would incorrectly transition to the state indicating the task of reaching the $\spadesuit$ in the left room. In such a scenario, even if the noise were to disappear completely after this tampering, the agent would be doomed to fail the task, as it would effectively act in pursuit of the wrong objective. Therefore, the precise timing required for carrying out such an attack, coupled with the random nature of the noise injected in our experiments, explains the low decrease in success rate observed in the SymbolWorld domain. Finally, the difference between the results obtained in the two domains highlights one key aspect of an RM-based agent's robustness to noise: its dependency on the environment at hand, in terms of both reward machine structure and state dynamics.

%% file: sections/conclusions.tex
\label{sec:conclusions}

Robustness to noise is of utmost importance in reinforcement learning systems, particularly in military contexts where high stakes and uncertain environments prevail.
Reward machines offer a powerful tool to express complex reward structures in RL tasks, enabling the design of tailored reinforcement signals that align with mission objectives. 
In this paper, we consider the problem of the robustness of intelligence-driven reinforcement learning based on reward machines. 
The results gathered, and, in particular, the one for CookieWorld \--- \cf \cref{tab:rescookie} \--- where noise and uncertainty play a bigger role, suggest the need for further research to harden current state-of-the-art reinforcement learning approaches prior to being mission-critical-ready. In future work, we plan to consider evidential learning and reasoning \cite{cerutti2022handling,cerutti2022evidential}  as possible techniques for ensuring robust exploitation of learned policies.